\documentclass[11pt]{article}
\usepackage{eamt23}
\usepackage{times}
\usepackage{url}
\usepackage{latexsym}
\usepackage[small,bf]{caption} 
\usepackage{tikz}
\usetikzlibrary{calc,matrix,arrows.meta,positioning}

\usepackage{xcolor}
\usepackage{multirow}
\usepackage{booktabs}

\let\citep\cite
\let\citet\newcite

\title{An Open-Source Gloss-Based Baseline \\ for Spoken to Signed Language Translation}

\author{
\parbox{0.65\linewidth}{\centering
Amit Moryossef$^{1,2}$, Mathias Müller$^2$, Anne Göhring$^2$, Zifan Jiang$^2$, Yoav Goldberg$^1$, Sarah Ebling$^2$    }
\\
$^1$Bar-Ilan University, 
$^2$University of Zurich
\\
\texttt{amitmoryossef@gmail.com}
\\\\
\url{https://github.com/ZurichNLP/spoken-to-signed-translation}}

\date{}

\begin{document}
\maketitle

\begin{abstract}

Sign language translation systems are complex and require many components. As a result, it is very hard to compare methods across publications. We present an open-source implementation of a text-to-gloss-to-pose-to-video pipeline approach, demonstrating conversion from German to Swiss German Sign Language, French to French Sign Language of Switzerland, and Italian to Italian Sign Language of Switzerland. 
We propose three different components for the text-to-gloss translation: a lemmatizer, a rule-based word reordering and dropping component, and a neural machine translation system. Gloss-to-pose conversion occurs using data from a lexicon for three different signed languages, with skeletal poses extracted from videos. To generate a sentence, the text-to-gloss system is first run, and the pose representations of the resulting signs are stitched together.
\end{abstract}

\section{Introduction}
Sign language plays a crucial role in communication for many deaf\footnote{We follow the recent convention of abandoning a distinction between ``Deaf'' and ``deaf'', using the latter term also to refer to (deaf) members of the sign language community \citep{kusters-et-al-2017,napier-leeson-2016}.} individuals worldwide. 
However, producing sign language content is often a challenging, laborious, and time-consuming process, requiring skilled translators/interpreters for effective communication. Recent technological advancements have led to the development of automated sign language translation systems, which have the potential to increase accessibility for the deaf community and enhance communication.

One of the critical issues in this field is the lack of a reproducible and reliable baseline for sign language translation systems. Without a baseline, it is challenging to measure the progress and effectiveness of new methods and systems. Additionally, the absence of such a baseline makes it difficult for new researchers to enter the field, hampers comparative evaluation, and discourages innovation. 

Addressing this gap, this paper presents an open-source implementation of a text-to-gloss-to-pose-to-video pipeline approach for sign language translation, extending the work of Stoll et al. \shortcite{stoll2018sign,stoll2020text2sign}. Our main contribution is the development of an open-source, reproducible baseline that can aid in making sign language translation systems more available and accessible, particularly in resource-limited settings. This open-source approach allows the community to identify issues, work together on improving these systems, and facilitates research into novel techniques and strategies for sign language translation

Our approach involves three alternatives for text-to-gloss translation, including a lemmatizer, a rule-based word reordering and dropping component, and a neural machine translation (NMT) system. For gloss-to-pose conversion, we use lexicon-acquired data for three signed languages, including Swiss German Sign Language (DSGS), Swiss French Sign Language (LSF-CH), and Swiss Italian Sign Language (LIS-CH). 
We extract skeletal poses using  a state-of-the-art pose estimation framework, and apply a series of improvements to the poses, including cropping, concatenation, and smoothing, before applying a smoothing filter.

\begin{figure*}
  \centering
  
  \resizebox{\linewidth}{!}{\input{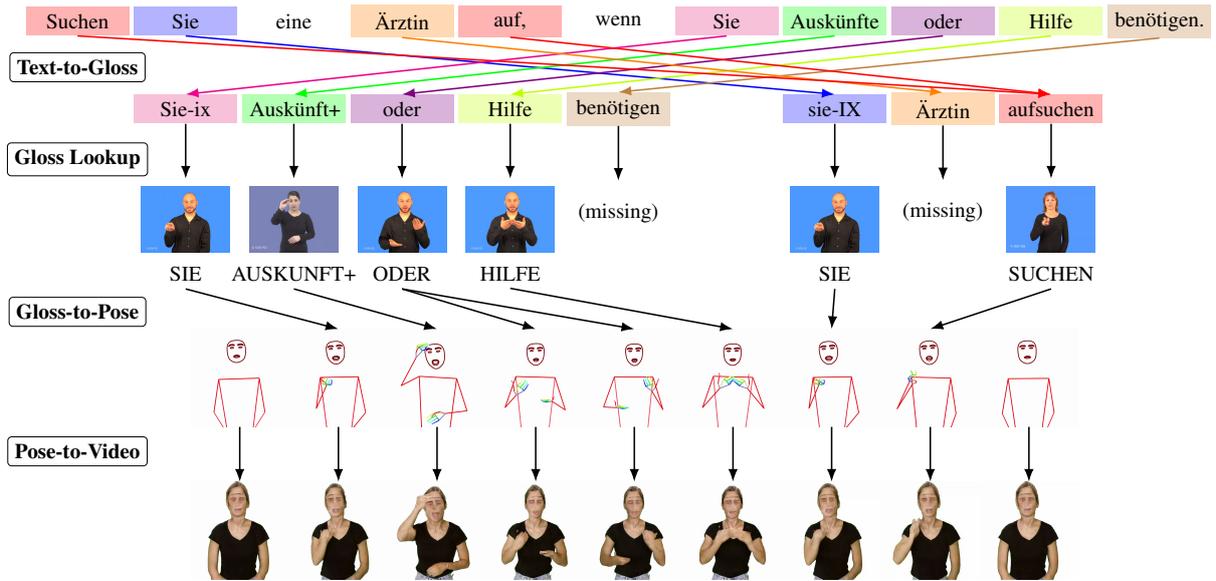}}
  \caption{The figure depicts the entire pipeline of the proposed text-to-gloss-to-pose-to-video approach for sign language translation. Starting with a German sentence, the system applies text-to-gloss translation, for example, using a rule-based word reordering and dropping component. The resulting gloss sequence is used to search for relevant videos from a lexicon of Swiss German Sign Language (DSGS). The poses of each relevant video are then extracted and concatenated in the gloss-to-pose step to create a pose sequence for the sentence, which is then transformed back to a (synthesized) video using the pose-to-video model.  The figure demonstrates the transformation of the sentence ``Suchen Sie eine Ärztin auf, wenn Sie Auskünfte oder Hilfe benötigen." (`Seek out a doctor if you need information or assistance.') to a sequence of glosses, the search for relevant videos for each gloss, the concatenation of pose videos, and the final video output.}
  \label{fig:pipeline}
\end{figure*}

\section{Background}

Sign language translation can be accomplished in various ways. In this section, we focus on the pipeline approach that involves text-to-gloss, gloss-to-pose, and, optionally, pose-to-video techniques. 
The text-to-gloss technique translates spoken language text into sign language glosses, which are then converted into a sequence of poses by gloss-to-pose techniques, and into a photorealistic video using pose-to-video techniques.

This pipeline offers the benefit of preserving the content of the sentence, while exhibiting a tendency for verbosity and a lower degree of fluency. 
In this section, we explore each of the pipeline components comprehensively and examine recent progress in sign language translation utilizing these methods.

\subsection{Text-to-Gloss}

Text-to-gloss, an instantiation of sign language translation, is the
task of translating between a spoken language text and sign language
glosses. It is an appealing area of research because of its simplicity for integrating in existing NMT pipelines, despite recent works such as \citet{yin-read-2020-better} and \citet{muller2022considerations} claim that glosses are an inefficient representation of sign language, and that glosses are not a complete representation of signs \citep{pizzuto:06001:sign-lang:lrec}. 

\citet{zhao2000machine} used a Tree Adjoining Grammar (TAG)-based system
to translate English sentences to American Sign Language (ASL) gloss
sequences. They parsed the English text and simultaneously assembled an
ASL gloss tree, using Synchronous TAGs
\citep{shieber1990synchronous,shieber1994restricting}, by associating
the ASL elementary trees with the English elementary trees and
associating the nodes at which subsequent substitutions or adjunctions
can occur. Synchronous TAGs have been used for machine translation
between spoken languages \citep{abeille1991using}, but this was the
first application to a signed language.

\citet{dataset:othman2012english} identified the need for a large
parallel sign language gloss and spoken language text corpus. They
developed a part-of-speech-based grammar to transform English sentences
from the Gutenberg Project ebooks collection \citep{lebert2008project}
into American Sign Language gloss. Their final corpus contains over 100
million synthetic sentences and 800 million words and is the most
extensive English-ASL gloss corpus we know of. Unfortunately, it is hard
to attest to the quality of the corpus, as the authors did not evaluate
their method on real English-ASL gloss pairs.

\citet{egea-gomez-etal-2021-syntax} presented a syntax-aware transformer for this task, 
by injecting word dependency tags to augment the embeddings inputted to the encoder. 
This involves minor modifications in the neural architecture leading to negligible impact on computational
complexity of the model.
Testing their model on the RWTH-PHOENIX-Weather-2014T \citep{cihan2018neural}, 
they demonstrated that injecting this additional information results in better translation quality. 

\subsection{Gloss-to-Pose}
Gloss-to-pose, subsumed under the task of sign language production, is
the task of producing a sequence of poses that adequately represent a
sequence of signs written as gloss.

To produce a sign language video, \citet{stoll2018sign} construct a
lookup table between glosses and sequences of 2D poses. They align all
pose sequences at the neck joint of a reference skeleton and group all
sequences belonging to the same gloss. Then, for each group, they apply
dynamic time warping and average out all sequences in the group to
construct the mean pose sequence. This approach suffers from not having
an accurate set of poses aligned to the gloss and from unnatural motion
transitions between glosses.

To alleviate the downsides of the previous work,
\citet{stoll2020text2sign} construct a lookup table of gloss to a group
of sequences of poses rather than creating a mean pose sequence. They
build a Motion Graph \citep{min2012motion}, which is a Markov process
used to generate new motion sequences that are representative of natural
motion, and select the motion primitives (sequence of poses) per gloss
with the highest transition probability. To smooth that sequence and
reduce unnatural motion, they use a Savitzky--Golay motion transition
smoothing filter \citep{savitzky1964smoothing}.

\subsection{Pose-to-Video}

Pose-to-video, also known as motion transfer or skeletal animation in
the field of robotics and animation, is the conversion of a sequence of
poses to a video. This task is the final ``rendering'' of sign language
in a visual modality.

\citet{pose:chan2019everybody} demonstrated a semi-supervised approach
where they took a set of videos, ran pose estimation with OpenPose
\citep{pose:cao2018openpose}, and learned an image-to-image translation
\citep{isola2017image} between the rendered skeleton and the original
video. They demonstrated their approach on human dancing, where they
could extract poses from a choreography and render any person as if
\emph{they} were dancing. They predicted two consecutive frames for
temporally coherent video results and introduced a separate pipeline for
a more realistic face synthesis, although still flawed.

\citet{pose:wang2018vid2vid} suggested a similar method using DensePose
\citep{pose:alp2018densepose} representations in addition to the
OpenPose \citep{pose:cao2018openpose} ones. They formalized a different
model, with various objectives to optimize for, such as
background-foreground separation and temporal coherence by using the
previous two timestamps in the input.

Using the method of \citet{pose:chan2019everybody} on ``Everybody Dance
Now'', \citet{pose:girocan2020slrtp} asked, ``Can Everybody Sign Now?''
and investigated if people could understand sign language from
automatically generated videos. They conducted a study in which
participants watched three types of videos: the original signing videos,
videos showing only poses (skeletons), and reconstructed videos with
realistic signing. The researchers evaluated the participants'
understanding after watching each type of video. The results of the
study revealed that participants preferred the reconstructed videos over
the skeleton videos. However, the standard video synthesis methods used
in the study were not effective enough for clear sign language
translation. Participants had trouble understanding the reconstructed
videos, suggesting that improvements are needed for better sign language
translation in the future.

As a direct response, \citet{saunders2020everybody} showed that like in
\citet{pose:chan2019everybody}, where an adversarial loss was added to
specifically generate the face, adding a similar loss to the hand
generation process yielded high-resolution, more photo-realistic
continuous sign language videos. To further improve the hand image
synthesis quality, they introduced a keypoint-based loss function to
avoid issues caused by motion blur.

In a follow-up paper, \citet{anonysign} introduced the task of Sign
Language Video Anonymisation (SLVA) as an automatic method to anonymize
the visual appearance of a sign language video while retaining the
original sign language content. Using a conditional variational
autoencoder framework, they first extracted pose information from the
source video to remove the original signer appearance, then generated a
photo-realistic sign language video of a novel appearance from the pose
sequence. The authors proposed a novel style loss that ensures style
consistency in the anonymized sign language videos.

\section{Method}

In this section, we provide an overview of our text-to-gloss-to-pose-to-video pipeline, detailing the components and how they work together to convert input spoken language text into a sign language video. 
The pipeline consists of three main components: text-to-gloss translation, gloss-to-pose conversion, and pose-to-video animation. 
For text-to-gloss translation, we provide three different alternatives: a lemmatizer, a rule-based word reordering and dropping component, and a neural machine translation system. Figure \ref{fig:pipeline} illustrates the entire pipeline and its components.

\subsection{Pipeline}
Below, we describe the high-level structure of our pipeline, including the text-to-gloss translation, gloss-to-pose conversion, and pose-to-video animation components:

\begin{enumerate}
\item \textbf{Text-to-Gloss Translation:} The input (spoken language) text is first processed by the text-to-gloss translation component, which converts it into a sequence of glosses. 

\item \textbf{Gloss-to-Pose Conversion:} The sequence of glosses generated from the previous step is then used to search for relevant videos from a lexicon of signed languages (e.g., DSGS, LSF-CH, LIS-CH). We extract the skeletal poses from the relevant videos using a state-of-the-art pre-trained pose estimation framework. These poses are then cropped, concatenated, and smoothed, creating a pose representation for the input sentence. 

\item \textbf{Pose-to-Video Generation:} The processed pose video is transformed back into a synthesized video using an image translation model, based on a custom training of Pix2Pix.
\end{enumerate}

\subsection{Implementation Details}
Our system accepts spoken language text as input and outputs an \emph{.mp4} video file, or a binary \emph{.pose} file, which can be handled by the \emph{pose-format} library \cite{moryossef2021pose-format} in Python and JavaScript. The \emph{.pose} file represents the sign language pose sequence generated from the input text. 
To make our system easy to use, we deploy it as an HTTP endpoint that receives text as input and outputs the \emph{.pose} file.
We provide a demonstration of our system using \url{https://sign.mt}, with support for the three signed languages of Switzerland.

We implement our pipeline using Python and package it using Flask, a lightweight web framework. This allows us to create an HTTP endpoint for our application, making it easy to integrate with other systems and web applications. Our system is deployed on a Google Cloud Platform (GCP) server, providing scalability and easy access. Furthermore, we release the source code of our implementation as open-source software, allowing others to build upon our work and contribute to improving the accessibility of sign language translation systems.

By implementing our system as an open-source Python application and deploying it as an HTTP endpoint, we aim to facilitate collaboration and improvements to sign language translation systems.

\section{Text-to-Gloss}

We explore three different components as part of text-to-gloss translation, including a lemmatizer (\S\ref{sec:lemmatizer}), a rule-based word reordering and dropping component (\S\ref{sec:rule-based}), and a neural machine translation (NMT) system (\S\ref{sec:nmt}).

\subsection{Lemmatizer}\label{sec:lemmatizer}

We use the \emph{Simplemma} simple multilingual lemmatizer for Python \citep{barbaresi_adrien_2023_7555188}. The lemmatizer reduces words to their base form (i.e., lemma), which is useful for our case, as it helps to preserve meaning while reducing the complexity of the input.
This approach is limited by the use of the simplistic context-free lemmatizer, since no sense information is captured in the lemma, which causes ambiguity.

\subsection{Word Reordering and Dropping}\label{sec:rule-based}

We generate near-glosses for sign language from spoken language text using a rule-based approach. The process from converting spoken language sentences into sign language gloss sequences can be naively summarized by a removal of word inflection, an omission of punctuation and specific words, and word reordering. 
To address these differences, we adopt the rule-based approach from \citet{moryossef-etal-2021-data} to generate near-glosses from spoken language: lemmatization of spoken words, PoS-dependent word deletion, and word order permutation. With their permission, we re-share these rules:

Specifically, we use spaCy \citep{ines_montani_2023_7553910} for lemmatization, PoS tagging and dependency parsing.
Unlike Simplelemma, the spaCy lemmatizer is language specific and context based.
We drop words that are not content words (e.g., articles, prepositions), as they are largely unused in signed languages, but keep possessive and personal pronouns as well as nouns, verbs, adjectives, adverbs, and numerals. We devise a short list of syntax transformation rules based on the grammar of the sign language and the corresponding spoken language. We identify the subject, verb, and object in the input text and reorder them to match the order used in the signed language. For example, for German-to-German Sign Language (\textit{Deutsche Gebärdensprache}, DGS), we reorder SVO sentences to SOV, move verb modifying adverbs and location words to the start of the sentence (a form of topicalization), move negation words to the end.

The specific rules we use for German to DGS/DSGS are:
\begin{enumerate}
    \item For each subject-verb-object triplet $(s,v,o) \in \mathcal{S}$, swap the positions of $v$ and $o$ in $\mathcal{S}$
    \item Keep all tokens $t \in \mathcal{S}$ if \textbf{PoS}$(t) \in$ \{noun, verb, adjective, adverb, numeral, pronoun\}
    \item If \textbf{PoS}$(t) =$ adverb and \textbf{HEAD}$(t) = $ verb, move $t$ to the start of $S$
    \item If \textbf{NER}$(t) =$ location, move $t$ to the start of $S$
    \item If \textbf{DEP}$(t) =$ negation, move $t$ to the end of $S$
    \item Lemmatize all tokens $t \in \mathcal{S}$
\end{enumerate}

We first split each sentence into separate clauses and reorder them before we apply these rules to each clause. Reordering the clauses may be needed for conditional sentences where the conditional subordinate clause should precede the main clause, as in ``if\dots then\dots''. These rules allow us to transform spoken language text into near-glosses that more closely match the word order and structure of sign language. 
Overall, our rule-based approach provides a flexible and effective way to generate near-glosses for sign language from spoken language text, with the ability to incorporate language-specific rules to capture the nuances of different sign languages. This approach employs a more accurate lemmatizer, however, it still suffers from word sense ambiguity.

\subsection{Neural Machine Translation}\label{sec:nmt}

As an alternative to rule-based transformations of text to glosses, we train a neural machine translation (NMT) system.

\paragraph{\textbf{Data}}

We use the Public DGS Corpus, a publicly available corpus of German Sign Language videos with annotated glosses \citep{dataset:hanke-etal-2020-extending}. Appendix \ref{appendix:corpus_specific_loading_and_gloss_preprocessing} explains our data loading and preprocessing in more detail.
We hold out a random sample of 1k training examples each for development and testing purposes. Table \ref{tbl:gloss_data} shows an overview of the number of sentence pairs in all splits.

\begin{table}[h]
    \begin{center}
    \small
    \begin{tabular}{lrrrr}
      \toprule
       \multicolumn{1}{c}{\textbf{Partition}} & \multicolumn{4}{c}{\textbf{Available Languages}} \\
      \midrule
        & \multicolumn{1}{r}{\textbf{EN}} & \multicolumn{1}{r}{\textbf{DGS$\cdot$DE}} & \multicolumn{1}{r}{\textbf{DGS$\cdot$EN}} & \multicolumn{1}{r}{\textbf{DE}} \\
      \midrule
       Train &  61912 & 61912 & 61912 & 61912 \\
       Dev &  1000 & 1000 & 1000 & 1000 \\
       Test &  1000 & 1000 & 1000 & 1000 \\
       \textbf{Total} &  \textbf{63912} & \textbf{63912} & \textbf{63912} & \textbf{63912} \\
      \bottomrule
    \end{tabular}
    \end{center}
    \caption{Number of sentence pairs used for gloss models. \\DGS$\cdot$DE=original gloss transcriptions, \\DGS$\cdot$EN=DGS glosses translated to English.}
    \label{tbl:gloss_data}
\end{table}

\paragraph{\textbf{Preprocessing}} 

Our preprocessing and model settings are inspired by OPUS-MT \citep{TiedemannThottingal:EAMT2020}. 
The only preprocessing step that we apply to all data is Sentencepiece segmentation \citep{kudo-2018-subword}. We learn a shared vocabulary with a desired total size of 1k pieces.

We additionally preprocess DGS glosses in a corpus-specific way, informed by the DGS Corpus glossing conventions \citep{konrad_reiner_2022_10251}. 
The exact steps are given in Appendix \ref{appendix:dgs_corpus}. See Table \ref{tbl:gloss_preprocessing_examples} for examples for this preprocessing step.
Overall the desired effect is to reduce the number of observed forms while not altering the meaning itself. 

\paragraph{\textbf{Core model settings}}

We train NMT models with Sockeye 3 \citep{hieber-etal-2022-sockeye3}. The models are standard Transformer models \citep{DBLP:journals/corr/VaswaniSPUJGKP17}, except with some hyperparameters modified for a low-resource scenario. E.g., dropout rate is set to a high value of 0.5 for all dropout layers of the model \citep{sennrich-zhang-2019-revisiting}.

The NMT system itself is trained with three-way weight tying between the source embeddings, target embeddings matrix and softmax output \citep{press-wolf-2017-using}.

We train a multilingual model, following the methodology described in \citet{johnson-etal-2017-googles} which inserts special tokens into all source sentences to indicate the desired target language. For comparison, we also train bilingual systems that can translate in only one direction each. Our automatic evaluation confirms that one multilingual system leads to higher translation quality than individual bilingual systems (see Appendix \ref{subsec:automatic_evaluation_of_translation_quality}).

\begin{table*}
    \begin{center}
    \small
    \begin{tabular}{lp{13cm}}
      \toprule
      \textbf{Before} & \texttt{\$INDEX1 ENDE1\^{} ANDERS1* SEHEN1 MÜNCHEN1B* BEREICH1A*} \\
      \textbf{After} & \texttt{\$INDEX1 ENDE1 ANDERS1 SEHEN1 MÜNCHEN1 BEREICH1} \\
      \midrule
      \textbf{Before} & \texttt{ICH1 ETWAS-PLANEN-UND-UMSETZEN1 SELBST1A* KLAPPT1* \$GEST-OFF\^{} BIS-JETZT1 GEWOHNHEIT1* \$GEST-OFF\^{}*} \\
      \textbf{After} & \texttt{ICH1 ETWAS-PLANEN-UND-UMSETZEN1 SELBST1 KLAPPT1 BIS-JETZT1 GEWOHNHEIT1} \\
      \bottomrule
    \end{tabular}
    \end{center}
    \caption{Examples for preprocessing of DGS glosses.}
    \label{tbl:gloss_preprocessing_examples}
\end{table*}

\subsection{Language Dependent Implementation}\label{sec:language-dependent}

In this paper, we study three sign languages: LIS-CH, LSF-CH and DSGS. For LIS-CH and LSF-CH we always apply our simple lemmatizer (\S\ref{sec:lemmatizer}) for the text-to-gloss step. The lemmatizer-only component is universally applicable to many more languages. However, it is worth noting that this approach does not capture the full spectrum of syntactic and morphological changes necessary in going from a spoken language to a sign language, which likely leads to suboptimal translations.

For DSGS, we explored different options for text-to-gloss, comparing the lemmatizer (\S\ref{sec:lemmatizer}), rule-based system (\S\ref{sec:rule-based}) and NMT system (\S\ref{sec:nmt}). We observed that the glosses output by the NMT system are less accurate than rule-based reordering. A potential explanation for this is that the system is trained on German Sign Language (DGS) data. Due to the inherent differences between DGS and DSGS, using the NMT system could result in inaccurate translations or out-of-lexicon glosses. Furthermore, we found that the NMT system is not robust to out-of-domain text or capitalization differences, which further limits its applicability in these scenarios.

In the end, for DSGS we opted to employ our rule-based system (\S\ref{sec:rule-based}), which has been tailored to accommodate the unique linguistic characteristics of DSGS, and produces the best results.

\section{Gloss-to-Pose}

Gloss-to-pose translation involves converting sign language glosses into a sequence of poses that adequately represent a sequence of signs.

We use the SignSuisse dataset \citep{SignSuisse}, which consists of sign language videos in three different languages. We extract skeletal poses from these videos using Mediapipe Holistic \citep{mediapipe2020holistic}, a state-of-the-art pose estimation framework that estimates 3D coordinates of various landmarks on the human body, including the face, hands, and body. We preprocess the poses by ensuring that the \texttt{body} wrists are in the same location as the \texttt{hand} wrists, removing the legs, hands, and face from the body pose, and cropping the videos in the beginning and end to avoid returning to a neutral body position.

We concatenate the poses for each gloss by finding the best `stitching' point that minimizes L2 distance. We then concatenate these poses, adding 0.2 seconds of `padding' in between, before applying cubic smoothing on each joint to ensure smooth transitions between signs, and filling in missing keypoints. Finally, we apply a Savitzky-Golay motion transition smoothing filter \citep{savitzky1964smoothing}, similar to \citet{stoll2020text2sign}, to reduce unnatural motion.

\section{Pose-to-Video}

We use a semi-realistic human-like avatar system to animate the poses generated by our approach. The avatar system is a Pix2Pix model \citep{pix2pix} adjusted to operate on pose sequences, not individual images. With her permission, we use the likeness of Maayan Gazuli\footnote{\url{https://nlp.biu.ac.il/~amit/datasets/GreenScreen/}}.
We use OpenCV \citep{opencv_library} to render the poses as images and feed them into the Pix2Pix model to generate realistic-looking video frames. The avatar system can run in real-time on supported devices and is integrated into \url{https://sign.mt} \citep{moryossef2023signmt}. 
This system is far from the state of the art, however, we believe that the open-source nature of it will bring rapid improvements, like faster inference speed, and higher animation quality.

\section{Future Work}

Here we include several future work directions that we believe have the potential to further enhance the performance and user experience of our system for text-to-gloss-to-pose-to-video generation, and we look forward to exploring these possibilities in the future, together with the open-source community.

\subsection{Qualitative Evaluation}
To evaluate the effectiveness of our approach, we will conduct a study to gather first impressions from deaf users. We already recruited a group of deaf individuals and will ask them to use our system to translate text into sign language videos.

Each participant will be asked to provide feedback on the system after using it to translate five different sentences from German into DSGS. We will provide the sentences to the participants, and they will be asked to sign the translations generated by our system. After each sentence, the participant will be asked to provide feedback on the accuracy of the translation, the quality of the poses and/or synthesized video, and the overall usability of the system.

\subsection{Gloss Sense Disambiguation}
The current approach to text-to-gloss translation relies on a simple lemmatizer and a rule-based word reordering and dropping component, which can lead to ambiguity in the glosses produced. In the future, we can enhance our system by incorporating gloss sense disambiguation to better capture the intended meaning of the input text. Our NMT approach responds with gloss IDs from the MeineDGS corpus, which already are sense-disambiguated. Annotation of our sign language lexicon with senses will allow us to retrieve the relevant sense.

\subsection{Handling Unknown Glosses}
Where we encounter a gloss that does not exist in our lexicon, we propose exploring alternative methods to generate a video for it. 
One possible solution is to leverage another lexicon that includes a written representation of the gloss in question (e.g., SignWriting \cite{writing:sutton1990lessons} or HamNoSys \cite{writing:prillwitz1990hamburg}), or to employ a neural machine translation system to translate the individual concept to a writing system. Utilizing the capabilities of machine translation to embed words, we can perform a fuzzy match, addressing issues such as synonyms.

Additionally, for named entities such as proper nouns and place names that are not covered by our current gloss-to-pose conversion system, we could revert to fingerspelling them.

Once we have the written representation, we can use a system like Ham2Pose \cite{shalevarkushin2023ham2pose} to generate a single sign video from the writing. When combined with fingerspelling for named entities, this approach should enable greater coverage of the language. 

\subsection{Handling Unknown Gloss Variations}

In situations where the required gloss variation is not present in the lexicon but a related gloss exists, we propose developing a system that can modify the known gloss to generate the desired variation. This would allow for better handling of unknown gloss variations and increase the accuracy of the information conveyed by the signing.

\subsubsection{Number Forms}
For words like \emph{KINDER} (children), we may encounter glosses such as \emph{KIND+}, which represent ``child" in plural form. Assuming that we have \emph{KIND} in our lexicon but not \emph{KINDER}, a system could be developed to modify signs to plural forms, such as by repeating movements or incorporating specific handshapes or locations that indicate plurality in the target sign language. Conversely, if we only have the plural form of a gloss in our lexicon, the system could be designed to generate the singular form by removing or modifying the elements that indicate plurality.

\subsubsection{Part of Speech Conversion}
Another challenge arises when nouns or verbs exist in the lexicon, but their counterparts do not. For instance, if \emph{HELFEN} (to help) is present in the dictionary as a verb, but \emph{HILFE} (help) does not exist as a noun, a system could be designed to modify signs from one part of speech to another, such as from verb to noun or noun to verb. This system could potentially involve morphological or movement modifications, depending on the linguistic rules of the target sign language.

\subsection{Post-editing Pose Sequences}
The current approach generates a sequence of poses that represent a sign language sentence. We believe that there is also room for improvement in terms of the fluency and naturalness of the generated sequence. Exploring the use of automatic post-editing techniques is necessary. 
One such approach could identify datasets that include sentences and gloss sequences, such as the Public DGS Corpus, then, using our gloss-to-pose approach generate a pose sequence with poses from the lexicon, and could learn a diffusion model between the synthetic and real pose sequences.

\section{Conclusions}

We presented an implementation of a text-to-gloss-to-pose-to-video pipeline for sign language translation, focusing on Swiss German Sign Language, Swiss French Sign Language, and Swiss Italian Sign Language. Our approach comprises three main components: text-to-gloss translation, gloss-to-pose conversion, and pose-to-video animation.

We explained the structure of our system and discussed its limitations, as well as future work directions to address them. These directions have the potential to improve our system, and we look forward to exploring them in collaboration with the open-source community.

The main contribution of this paper is the creation of a reproducible baseline for spoken to signed language translation. The system should serve as a baseline for comparison with more sophisticated sign language translation systems and can be improved upon by the community. You can try our system for the three signed languages of Switzerland on \url{https://sign.mt}.

\section*{Acknowledgements}

This work was funded by the EU Horizon 2020 project EASIER (grant agreement no. 101016982), the Swiss Innovation Agency  (Innosuisse) flagship IICT (PFFS-21-47), and the EU Horizon 2020 project iEXTRACT (grant agreement no. 802774).

\bibliography{references}
\bibliographystyle{eamt23}

\onecolumn
\newpage

\appendix

\section{SacreBLEU Signatures}
\label{appendix:sacrebleu_signatures}

\begin{table*}[!ht]
    \centering
    \scriptsize
    \begin{tabular}{lp{10cm}}
      \toprule
      \textbf{BLEU with internal tokenization} & \texttt{BLEU+case.mixed+numrefs.1+smooth.exp+tok.13a+version.1.4.14} \\
      \midrule
      \textbf{BLEU without internal tokenization} & \texttt{BLEU+case.mixed+numrefs.1+smooth.exp+tok.none+version.1.4.14} \\
      \midrule
      \textbf{CHRF} & \texttt{chrF2+numchars.6+space.false+version.1.4.14} \\
      \bottomrule
    \end{tabular}
    
    \caption{SacreBLEU signatures for evaluation metrics.}
    \label{tbl:gloss_sacrebleu_signatures}
\end{table*}

\section{Corpus-specific Loading and Gloss Preprocessing}
\label{appendix:corpus_specific_loading_and_gloss_preprocessing}

In general, we provide tools to automatically download all relevant examples from the corpus websites and only keep examples that have both a spoken language translation and a gloss transcription.
We experiment with corpus-specific preprocessing for glosses, informed by sign language linguistics and the glossing conventions of the corpora.

\subsection{DGS Corpus}
\label{appendix:dgs_corpus}
We download and process release 3.0 of the corpus.
To DGS glosses we apply the following modifications derived from the DGS Corpus transcription conventions \citep{konrad_reiner_2022_10251}:

\begin{itemize}
    \item Removing entirely two specific gloss types that cannot possibly help the translation: \texttt{\$GEST-OFF} and \texttt{\$\$EXTRA-LING-MAN}.
    \item Removing \textit{ad-hoc} deviations from citation forms, marked by \texttt{*}. Example: \texttt{ANDERS1*} $\rightarrow$ \texttt{ANDERS1}.
    \item Removing the distinction between type glosses and subtype glosses, marked by \texttt{\^{}}. Example: \texttt{WISSEN2B\^{}} $\rightarrow$ \texttt{WISSEN2B}. 
    \item Collapsing phonological variations of the same type that are meaning-equivalent. Such variants are marked with uppercase letter suffixes. Example: \texttt{WISSEN2B} $\rightarrow$ \texttt{WISSEN2}.
    \item Deliberately keep numerals (\texttt{\$NUM}), list glosses (\texttt{\$LIST}) and finger alphabet (\texttt{\$ALPHA}) intact, except for removing handshape variants.
\end{itemize}

See Table \ref{tbl:gloss_preprocessing_examples} for examples for this preprocessing step. Overall these simplifications should reduce the number of observed forms while not affecting the machine translation task. For other purposes such as linguistic analysis our preprocessing would of course be detrimental.

\subsection{Evaluation: Text-to-Gloss NMT}
\label{subsec:automatic_evaluation_of_translation_quality}

We perform an automatic evaluation of translation quality. We measure translation quality with BLEU \citep{papineni-etal-2002-bleu} and CHRF \citep{popovic-2016-chrf}, computed with the tool SacreBLEU \citep{post-2018-call}.
See Table \ref{tbl:gloss_sacrebleu_signatures} in Appendix \ref{appendix:sacrebleu_signatures} for all SacreBLEU signatures.

Whenever gloss output is evaluated we disable BLEU's internal tokenization, as advocated by \citet{muller2022considerations}. Earlier works did not consider this detail and therefore our BLEU scores may appear low in comparison.

Finally, because DGS glosses are preprocessed in a corpus-specific way (see above), they are evaluated against a preprocessed reference as well, since this process cannot be reversed after translation. This means that corpus-specific preprocessing for DGS glosses simplifies the translation task overall, compared to a system that predicts glosses in their original forms.

Table \ref{tbl:nmt_evaluation_results} reports the translation quality of our machine translation systems, as measured by CHRF. The table shows that one multilingual system that can translate between DGS and German leads to higher translation quality than two bilingual systems.

\begin{table}
    \begin{center}
    \footnotesize
    \begin{tabular}{lcc}
      \toprule
 & \multicolumn{1}{c}{\textbf{DGS$\rightarrow$DE}} & \multicolumn{1}{c}{\textbf{DE$\rightarrow$DGS}} \\
      \midrule
      Bilingual & 28.610 & - \\
      Bilingual & - & 32.920 \\
      \midrule
      Multilingual: all DE and DGS directions & 28.210 & 34.760 \\
      \bottomrule
    \end{tabular}
    \end{center}
    \caption{CHRF scores of the multilingual translation system compared to bilingual systems.}
    \label{tbl:nmt_evaluation_results}
\end{table}

\end{document}